\def\year{2019}\relax
\documentclass[letterpaper]{article} 
\usepackage{aaai}  
\usepackage{times}  
\usepackage{helvet} 
\usepackage{courier}  
\usepackage[hyphens]{url}  
\usepackage{graphicx} 
\usepackage{graphicx}  
\usepackage{caption}
\usepackage{subcaption}
\title{A Neural Architecture for Person Ontology population}
\author{Balaji Ganesan,\textsuperscript{\rm 1} Riddhiman Dasgupta,\textsuperscript{\rm 1}\thanks{Work done while the author was at IBM Research, India} Akshay Parekh,\textsuperscript{\rm 2} Hima Patel,\textsuperscript{\rm 1} Berthold Reinwald \textsuperscript{\rm 3}\\
IBM Research, India,\textsuperscript{\rm 1} IIT Guwahati, India,\textsuperscript{\rm 2} IBM Research, Almaden\textsuperscript{\rm 3}\\
\{bganesa1, himapatel\}@in.ibm.com, \ ridasgu@microsoft.com, \ akshayparakh@iitg.ac.in,
\ reinwald@us.ibm.com
}
 
\begin{document}

\maketitle

\begin{abstract}
A person ontology comprising concepts, attributes and relationships of people has a number of applications in data protection, de-identification, population of knowledge graphs for business intelligence and fraud prevention. While artificial neural networks have led to improvements in Entity Recognition, Entity Classification, and Relation Extraction, creating an ontology largely remains a manual process, because it requires a fixed set of semantic relations between concepts. In this work, we present a system for automatically populating a person ontology graph from unstructured data using neural models for Entity Classification and Relation Extraction. We introduce a new dataset for these tasks and discuss our results.
\end{abstract}

\section{Introduction}
We can define Personal Data Entity (PDE) as any information about a person. Such information can be present in both the public domain as well as in personal data. 

\begin{figure}[htb]
    \centering
    \includegraphics[width=\linewidth]{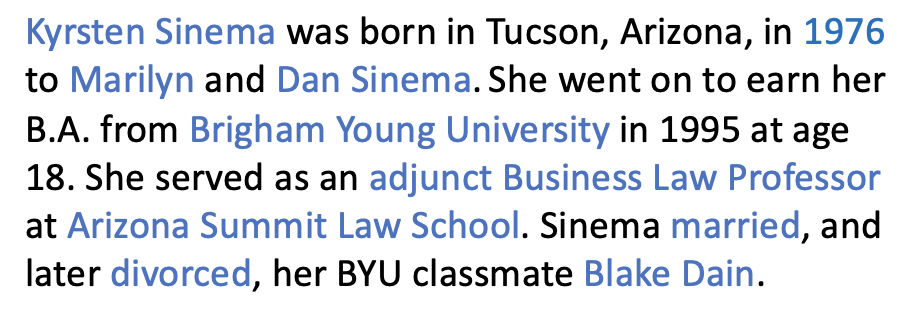}
    \caption{Personal Data Entities in unstructured text.}
    \label{fig:relation_example}
\end{figure}{}

The above sentences are from the publicly available Wikipedia page of an elected official. These sentence by themselves cannot be considered as personal data. But they contain Personal Data Entities (PDEs), i.e. entities which describe a person. A news article may also contain such mentions about an elected official.

For a number of applications in Data Protection, fraud prevention and business intelligence, there is a need to extract Personal Data Entities, classify them at a fine grained level, and identify relationships between people. Manually created Person Ontologies are used for this purpose in many enterprises cutting across domains. The challenges however are in populating an Ontology Graph based on such an Ontology.

\begin{figure}[htb]
    \centering
    \includegraphics[width=0.9\linewidth]{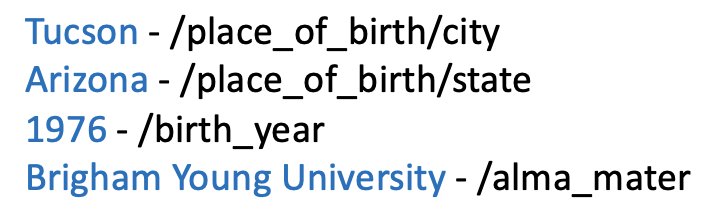}
    \caption{Attributes of Person Entity}
    \label{fig:fine_grained_types}
\end{figure}{}

The first challenge in Ontology population is in identifying attributes at a fine grained level. In Figure \ref{fig:relation_example}, \textit{Brigham Young University} could be classified coarsely as ORGANISATION by a Named Entity Recognizer. In recent years, a number of Neural Fine Grained Entity Classification (NFGEC) models have been proposed, which assign fine grained labels to entities based on context. They could might type \textit{Brigham Young University} as \slash{org}\slash{education}. However the focus of such systems has not been on PDEs. They do not treat the problem of identifying PDEs any different from other entities. For the purpose of Ontology population, it might be desirable to assign the below labels.

In typical relation extraction tasks, a person and their place of birth could be considered a relation. However in a Person Ontology, we might want to have only people as a first class concept. Hence we want to extract relations between people, but other entities like place of birth could be considered attributes of the Person entity. In our example, we might be satisfied with the following relations, though class mate, ex-spouse will be more accurate.

\begin{figure}[htb]
    \centering
    \includegraphics[width=0.9\linewidth]{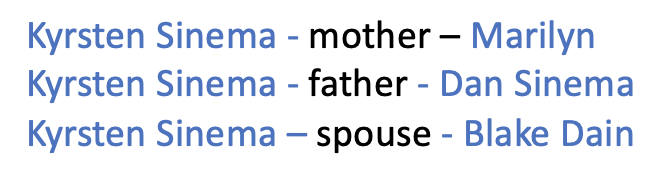}
    \caption{Relations among Person Entities}
    \label{fig:relations}
\end{figure}{}

We summarize our contributions in this work as follows:
\begin{itemize}
    \item We introduce a new dataset annotated with 36 Personal Data Entity Types (PDET) and 9 Personal Data Entity Relations (PDER).
    \item We propose an approach to improve state of the art models for fine-grained entity classification, only using light weight features.
    \item We share our results on running a semantic relation model on sentences rather than triples, by incorporating sentence embedding. These results however have not been encouraging.
    \item Finally, we implement a personal data ontology population pipeline by using graph neural networks to augment the relations from the relation extraction model.
\end{itemize}
    
\section{Related Work}
\subsection{Entity Classification}
Entity classification is a well known research problem in Natural Language Processing (NLP).~\cite{ling2012fine} proposed the FIGER system for fine grained entity recognition. In recent years, ~\cite{yogatama2015embedding}, ~\cite{shimaoka2017neural}, ~\cite{choi2018ultra} have proposed different neural models for context dependent fine grained entity classification. ~\cite{abhishek2017fine}, ~\cite{xu2018neural} proposed improvements to such models using better loss functions. ~\cite{yogatama2015embedding} showed the relevance of hand-crafted features for entity classification.~ \cite{shimaoka2017neural} further showed that entity classification performance varies significantly based on the input dataset (more than usually expected in other NLP tasks).

\subsection{Relation Extraction}

Models making use of dependency parses of the input sentences, or dependency-based models, have proven to be very effective in relation extraction, as they can easily capture long-range syntactic relations. \cite{zhang2018graph} proposed an extension of graph convolutional network that is tailored for relation extraction. Their model encodes the dependency structure over the input sentence with efficient graph convolution operations, then extracts entity-centric representations to make robust relation predictions. Hierarchical relation embedding (HRE) focuses on the latent hierarchical structure from the data. \cite{chen2018on2vec} introduced neighbourhood constraints in node-proximity-based or translational methods.

\subsection{Datasets}
\cite{ling2012fine} introduced the  Wiki dataset that consists of 1.5M
sentences  sampled  from  Wikipedia  articles. OntoNotes dataset by~\cite{weischedel2013ontonotes} consists  of 13,109  news  documents  where  77  test  documents are  manually  annotated~\cite{gillick2014context}. BBN dataset by~\cite{weischedel2005bbn} consists of 2,311  Wall  Street Journal  articles  which  are  manually  annotated  using  93  types.~\cite{murty2017finer} have proposed a much larger label set based on Freebase. 

\section{Personal Data Ontology}

\begin{figure}
    \centering
    \includegraphics[width=0.7\columnwidth]{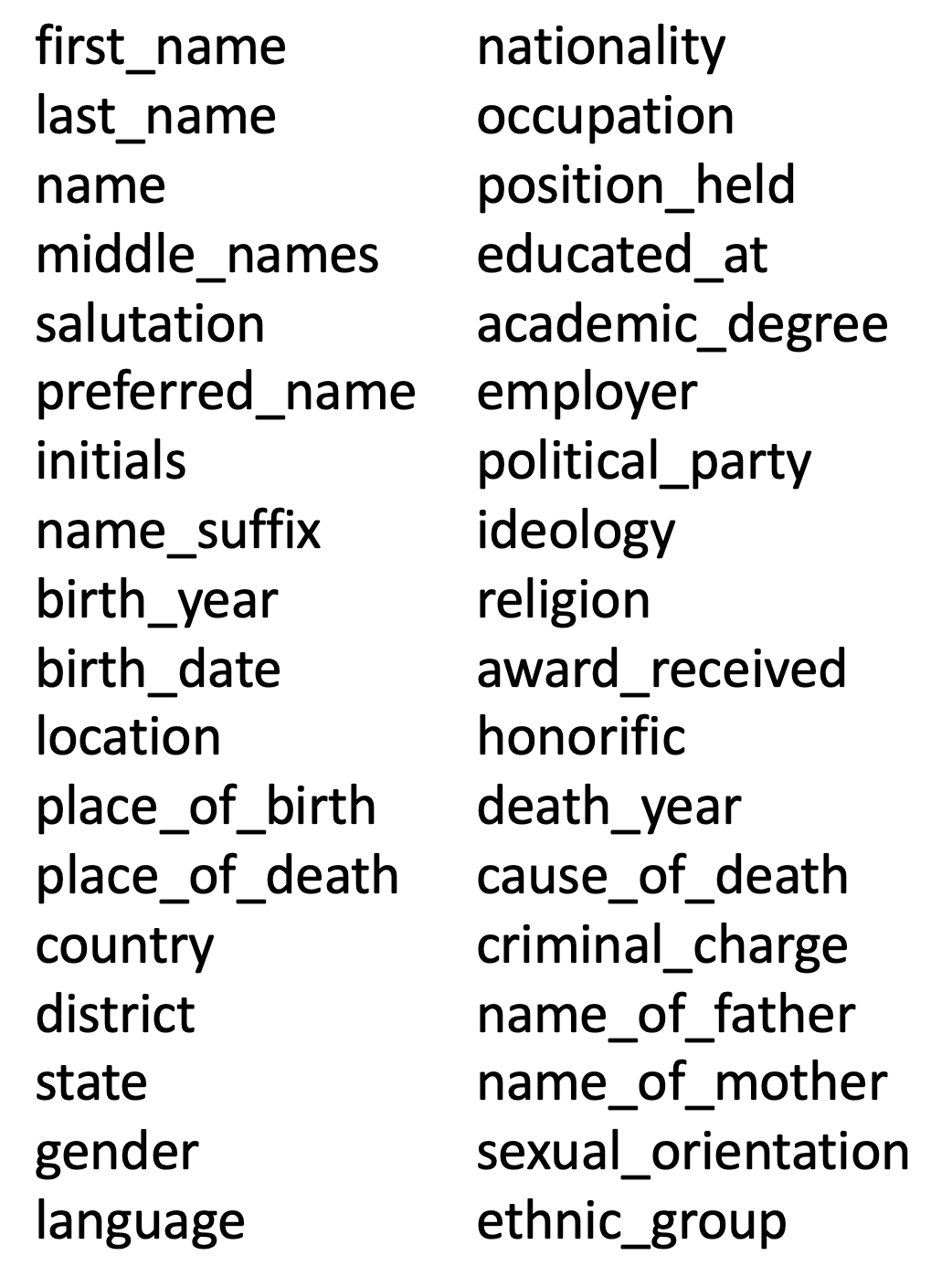}
    \caption{Personal Data Entity Types (PDET)}
    \label{fig:pde_hierarchy} 
\end{figure}
\begin{figure}
    \centering
    \includegraphics[width=\columnwidth]{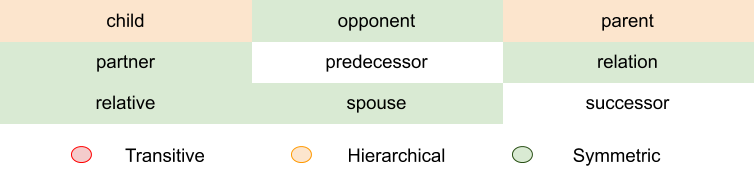}
    \caption{Personal Data Entity Relations (PDER)}
    \label{fig:pde_relations}
\end{figure}
\begin{figure}
    \centering
    \includegraphics[width=\columnwidth]{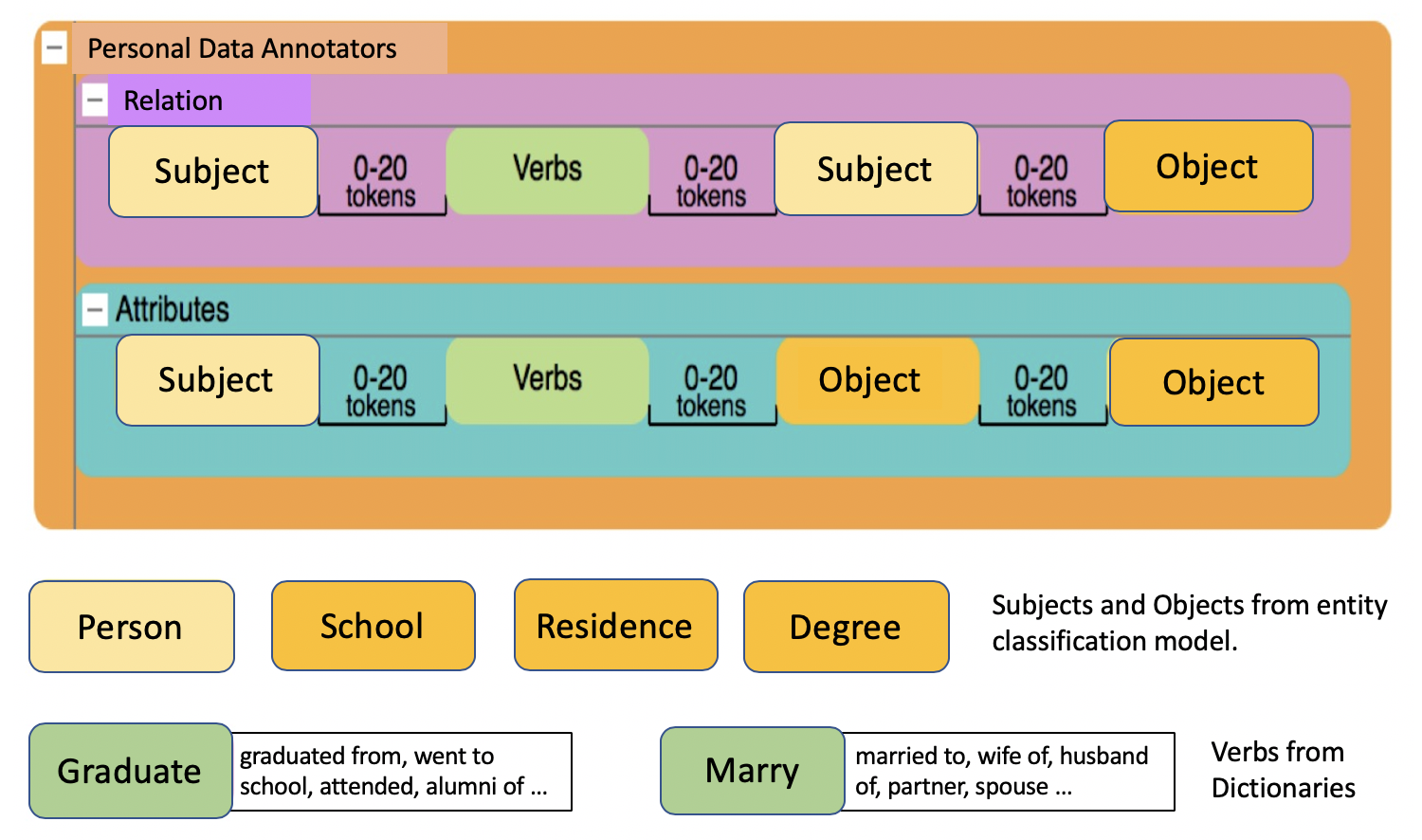}
    \caption{Personal Data Annotators}
    \label{fig:relation_annotators}
\end{figure}

\cite{ling2012fine} proposed the FIGER entity type hierarchy with 112 types.~\cite{gillick2014context} proposed the Google Fine Type (GFT) hierarchy and annotated 12,017 entity mentions with a total of 89 types from their label set. These two hierarchies are general purpose labels covering a wide variety of domains. \cite{dasgupta2018fine} proposed a larger set of Personal Data Entity Types with 134 entity types as shown in Figure~\ref{fig:pde_hierarchy}. We have selected the 36 personal data entity types, as shown in \ref{fig:pde_hierarchy} that were found in our input corpus.

For relation extraction labelset, YAGO \cite{yago} contained 17 relations, TACRED \cite{zhang2017position} proposed 41 relations and UDBMS (DBPedia Person) dataset \cite{UDMS_dataset} proposed 9 relations. We have used the 9 Personal Data Entity Relations (PDER) as shown in Figure-\ref{fig:pde_relations}.

\subsection{Personal Data Annotators}

Any system that assigns a label to a span of text can be called an annotator. In our case, these annotators assign an entity type to every entity mention. We have experimented with an enterprise (rule/pattern based) annotation system called SystemT introduced by \cite{chiticariu2010systemt}.

SystemT provides about 25 labels, which are predominantly coarse grained labels.

We use these personal data annotators in 3 ways:
\begin{itemize}
\item To annotate the dataset with entities for the entity classification task.     \item As part of the Personal Data Classification pipeline, where for some of the classes, the output of these PDAs are directly used as entity types. These are types like email address, zip codes, number where rule-based systems provide coarse labels at high precision.
\item To create a series of labeling functions that annotate relations between entities. These relations are used to bootstrap link prediction models, which in turn populate the Ontology Graph.
\end{itemize}

While neural networks have recently improved the performance of entity classification on general entity mentions, pattern matching and dictionary based systems continue to be used for identifying personal data entities in the industry.

We believe our proposed approach, consisting of modifications to state-of-the-art neural networks, will work on personal datasets for two reasons.~\cite{yogatama2015embedding} showed that hand-crafted features help, and~\cite{shimaoka2017neural} have shown that performance varies based on training data domain. We have incorporated these observations into our model, by using coarse types from rule-based annotators as side information. 

We used our Personal Data Annotators to create a number of labeling functions like those shown below to create a set of relations between the entities.

We have created this dataset from the Wikipedia page of US House of Representatives and the Members of the European Parliament. We obtained the names of 1196 elected representatives from the listings of these legislatures. These listings provide the names of the elected representatives and other details like contact information. However this semi-structured data by itself cannot be used for training a neural model on unstructured data. 

Hence, we first obtained the Wikipedia pages of elected representatives. We then used Stanford OpenNLP to split the text into sentences and tokenize the sentences. We ran the Personal Data Annotators on these sentences, providing the bulk of the annotations that are reported in Table~\ref{tab:datasets}.

We then manually annotated about 300 entity mentions which require fine grained types like \slash{profession}. The semi-structured data obtained from the legislatures had name, date of birth, and other entity mentions. We needed a method to find these entity mentions in the wikipedia text, and assign their column names or manual label as PDEs.

We used the method described in~\cite{chiticariu2010systemt} to identify the span of the above entity mentions in wikipedia pages. This method requires creation of dictionaries each named after the entity type, and populated with entity mentions. 
This approach does not take the context of the entity mentions while assigning labels and hence the data is somewhat noisy. However, labels for name, email address, location, website do not suffer much from the lack of context and hence we went ahead and annotated them.

\begin{table}[!htb]
    \begin{center}
    \begin{tabular}{llll}
        \textbf{}                       & \textbf{Elected Reps} \\
        \hline
        \textbf{Documents}               & 1196                 \\
        \textbf{Sentences}               & 38710                \\
        \textbf{Entity mentions}         & 211647               \\
        \textbf{Unique entity mentions}  & 45686                \\
        \textbf{Unique entity types}     & 91                   
    \end{tabular}
    \caption{Statistics on personal data annotations}
    \label{tab:datasets}
    \end{center}
\end{table}

\section{Neural Fine Grained Entity Classification}

\begin{figure*}
\centering
\begin{subfigure}{.5\textwidth}
  \centering
  \includegraphics[width=\linewidth]{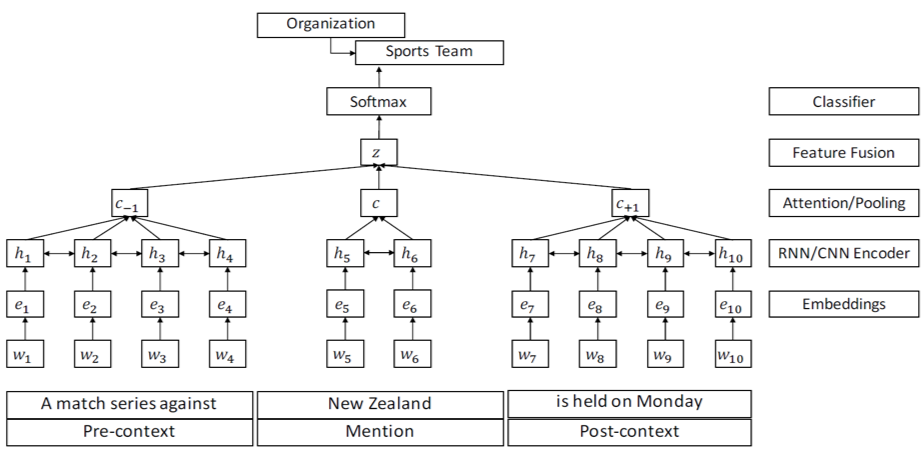}
  \caption{Entity Classification}
  \label{fig:nfgec_model}
\end{subfigure}%
\begin{subfigure}{.5\textwidth}
  \centering
  \includegraphics[width=\linewidth]{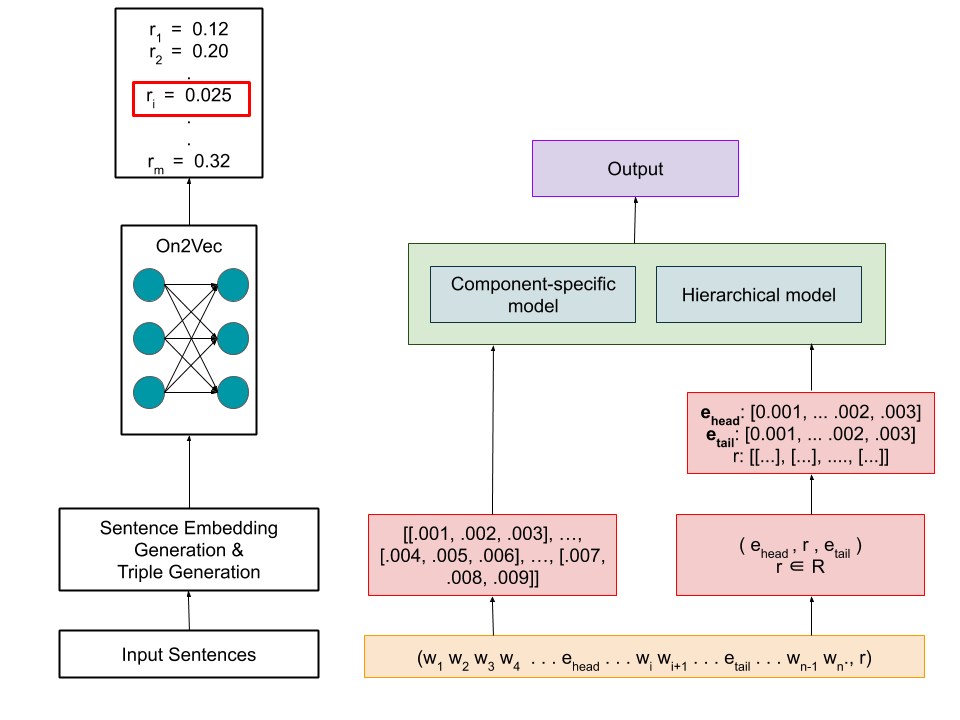}
  \caption{Relation Extraction}
  \label{fig:block}
\end{subfigure}
\caption{Neural Models for Entity Classification and Relation Extraction}
\label{fig:test}
\end{figure*}

We use the architecture described in \cite{dasgupta2018fine}, which in turn was based on ~\cite{shimaoka2017neural}. It consists of an encoder for the left and right contexts of the entity mention, another encoder for the entity mention itself, and a logistic regression classifier working on the features from the aforementioned encoders. An illustration of the model is shown in Figure~\ref{fig:nfgec_model}.

The major drawback of the features used in~\cite{shimaoka2017neural} was the use of custom hand crafted features, tailored for the specific task, which makes generalization and transferability to other datasets and similar tasks difficult. Building on these ideas, we have attempted to augment neural network based models with low level linguistic features which are obtained cheaply to push overall performance. Below, we elaborate on some of the architectural tweaks we attempt on the base model. 


Similar to~\cite{shimaoka2017neural}, we use two separate encoders for the entity mention and the left and right contexts. For the entity mention, we resort to using the average of the word embeddings for each word. For the left and right contexts, we employ the three different encoders mentioned in~\cite{shimaoka2017neural}, viz. 
\begin{itemize}
    \item The averaging encoder, which like the mention encoder, and uses the average as the context representation
    \item The RNN encoder, which runs an RNN over the context and takes the final state as the representation of the context
    \item The attentive encoder, which runs a bidirectional RNN over the context, and employs self-attention to obtain scores for each word, which are in turn used to get a weighted sum of the states to use as the representation.
\end{itemize}

Details of the different encoders can be found in~\cite{shimaoka2017neural}, and we omit them here for brevity. The features from the mention encoder, and the left and right context encoders are concatenated, and passed to a logistic regression classifier. If we consider $v_{left}$ to be the representation of the left context, $v_{right}$ to be the representation of the right context, and $v_{entity}$ to be the representation of the entity mention, each being $D$ dimensional then these features are concatenated to form $v = [ v_{left}, v_{right}, v_{entity}$, which is passed to the logistic regression classifier, which in turn computes the function:
\begin{equation}
    y = \frac{1}{1 + exp \left ( -W_y v \right )}
\end{equation}
where $W_y$ is the set of weights that project the features from a $3 \times D$ dimensional feature space to a $K$ dimensional output, where $K$ is the number of labels, and $0 \leq y_k \leq 1 \forall k \in K$. Since the output is a binary vector, we employ a binary cross entropy loss during training. Given the predictions $y$ and the ground truth $t$ for a sample, the loss is defined as:
\begin{equation}
    L(y, t) = \sum_{k=1}^{K} -t_k log(y_k) - (1-t_k)log(1-y_k)
\end{equation}
We employ stochastic mini-batch gradient descent to optimize the above loss function, and the details are specified later in the experimental results section.

\begin{table}[!htb]
    \begin{center}
    \begin{tabular}{llll}
        \textbf{Dataset}                 & \textbf{\# Test samples} & \textbf{\# Labels} \\
        \hline
        \textbf{OntoNotes}               & 8963                     & 89 \\
        \textbf{Elected Reps} & 16805                    & 91 
    \end{tabular}
    \caption{Statistics of the datasets used in our experiments.}
    \label{tab:dataset_details}
    \end{center}
\end{table}

The results on Elected Reps dataset as can be seen in Table ~\ref{tab:ontonotes_results}, clearly show the same trend, i.e. adding token level features improve performance across the board, for all metrics, as well as for any choice of encoder. The important thing to note is that these token level features can be obtained cheaply, using off-the-shelf NLP tools to deliver linguistic features such as POS tags, or using existing rule based systems to deliver task or domain specific type tags. This is in contrast to previous work such as~\cite{ling2012fine},~\cite{yogatama2015embedding} and others, who resort to carefully hand crafted features.

\begin{table}[!htb]
    \begin{center}
    \begin{tabular}{cccc}
    \hline
    \textbf{Dataset} & \textbf{Model} & \textbf{Macro F1} & \textbf{Micro F1} \\
    \hline
    {\textbf{OntoNotes}}
        & NFGEC      & 0.678    & 0.617 \\
        & NFGEC+       & 0.740    & 0.672 \\
        \cline{2-4}
    \hline
    {\textbf{Elected Reps}}  &   NFGEC  & 0.959 & 0.955  \\
                    &   NFGEC+ & 0.989 & 0.985  \\
                        \cline{2-4}
    \hline
    \end{tabular}
    \caption{Entity Classification performance with and without light weight features.}
    \label{tab:ontonotes_results}
    \end{center}
\end{table}

\section{Relation Extraction}

\begin{figure*}[htb]
    \includegraphics[width=\linewidth]{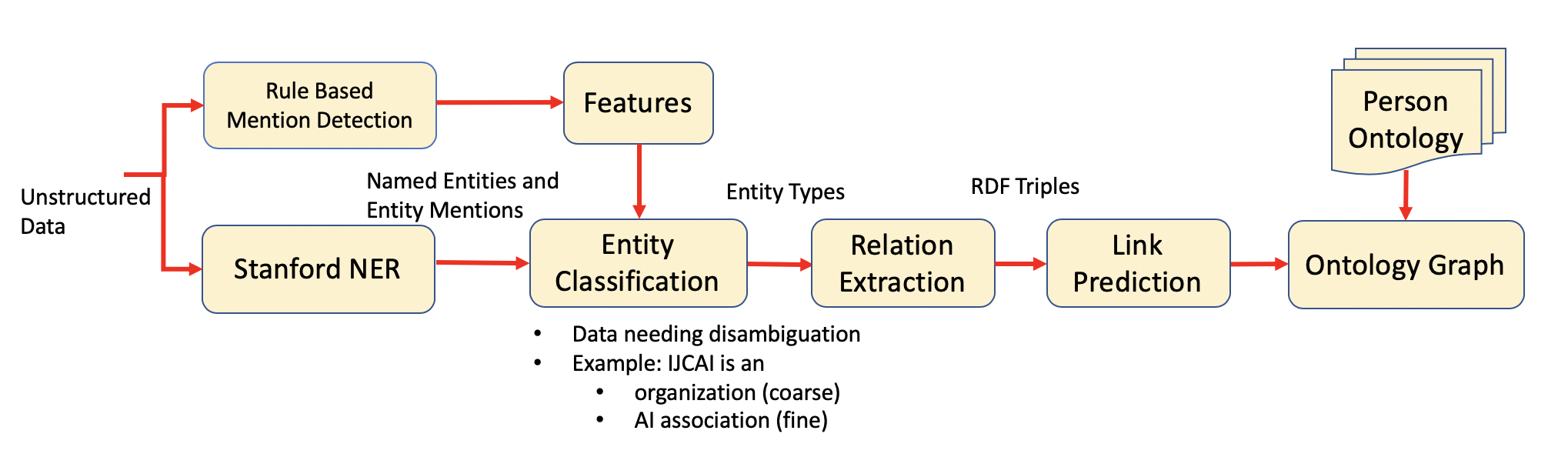}
    \caption{Person Ontology population pipeline}
    \label{fig:pipeline}
\end{figure*}
Extracting meaningful information from text requires models capable of extracting semantic relations, which are more comprehensive in terms of relational properties. 
A practical solution is to use translation-based graph embedding methods along with sentence embeddings, which will extract relation using vector representation of graph and sentence together. Where graph embedding will provide relation specific projection and sentence embedding will provide contextual information.
    
Majority of translation-based knowledge graph embedding methods project \textit{source} and \textit{target} entities in a k-dimensional vector. They typically focus only on simple relations, and less on comprehensive relations. In this work, we evaluate On2Vec \cite{chen2018on2vec} on the task of relation extraction. On2Vec proposed a two-component model, the \textit{Component Specific Model (CSM)}: encodes concepts and relation into low-dimensional embedding space without the loss of relational properties, such as symmetric and transitive. And the \textit{Hierarchical Model (HM)}: for better learning of hierarchical relations.
    
For generating sentence embedding, we use Universal Sentence Encoder (USE) \cite{cer2018universal} that makes use of Deep Averaging Network (DAN), whereby input embedding for words and bi-grams are averaged and passed to deep neural network to produce sentence embeddings. We tried a modification of On2Vec model, where we pass triples and sentence embedding generated using USE to the On2Vec's CSM (\ref{fig:block}). Sentence embedding is added to the energy function of CSM model, to provide the textual context.

We evaluate On2Vec model on YAGO60K, YAGO15K, TACRED and UDBMS (DBPedia Person) datasets. YAGO and UDBMS datasets contain triples, whereas TACRED contains sentences. We have mapped the relations in TACRED and YAGO datasets as transitive, hierarchical and symmetric. The results of our experiments are shown in Table \ref{tab:on2vec}. We observe that incorporating the sentence embedding is not helping the model.
    
    

\begin{table}[!htb]
    \begin{center}
    \begin{tabular}{cccc}
    \hline
    \textbf{Dataset} & \textbf{Input} & \textbf{Model} & \textbf{Accuracy} \\ \hline
    YAGO60K       & Triples  & On2Vec & 88.74 \\
    YAGO15K      & Triples  & On2Vec & 88.75 \\
    TACRED & Sentences & TACRED & 56.84 \\
    TACRED & Sentences & On2Vec+ & 17.8 \\
    UDBMS & Triples & On2Vec+ & 48.02 \\ \hline
    \end{tabular}
    \end{center}
    \caption{Performance of Relation Extraction models}
    \label{tab:on2vec}
\end{table}

\section{Person Ontology population pipeline}

We have implemented a pipeline for Personal Ontology population as shown in Figure~\ref{fig:pipeline}. This pipeline consists of existing personal data annotators, Stanford Named Entity Recognizer which provide rule based entity and relation extraction. We have then improved two state of the art models for entity classification and relation extraction as described in the previous sections. Finally we use a graph neural network for Link Prediction to infer more relationships between people mentioned in the corpus.

The input to our pipeline are text sentences. The outputs are person entities, their personal data as attributes and semantically rich relations between person entities. These can be used to populate a graph database like the one provided by networkx \cite{hagberg2008exploring}.

We present the results from training two graph neural networks on the Personal Data Entity (PDE) data extracted using our method and a similar DBPedia Person data which has been annotated by wikipedia users.

\begin{table}[!htb]
    \begin{center}
    \begin{tabular}{ccccccc}
    \hline
    \textbf{Dataset} & \textbf{Model} & \textbf{ROC AUC} & \textbf{Std. Dev.} \\
    \hline
    {\textbf{UDBMS}}  &   \textbf{GCN} & 0.4689 & 0.0280   \\
                    &   \textbf{P-GNN} & 0.6456 & 0.0185   \\
                        \cline{2-4}
    {\textbf{Elected Reps}}  &   \textbf{GCN} & 0.4047  & 0.09184  \\
                    &   \textbf{P-GNN} & 0.6473 & 0.02116   \\
    \hline
    \end{tabular}
    \caption{Comparison of Link Prediction on the UDBMS and Elected Representatives datasets}
    \label{tab:linkprediction}
    \end{center}
\end{table}

\begin{figure}[ht]
    \includegraphics[width=\columnwidth]{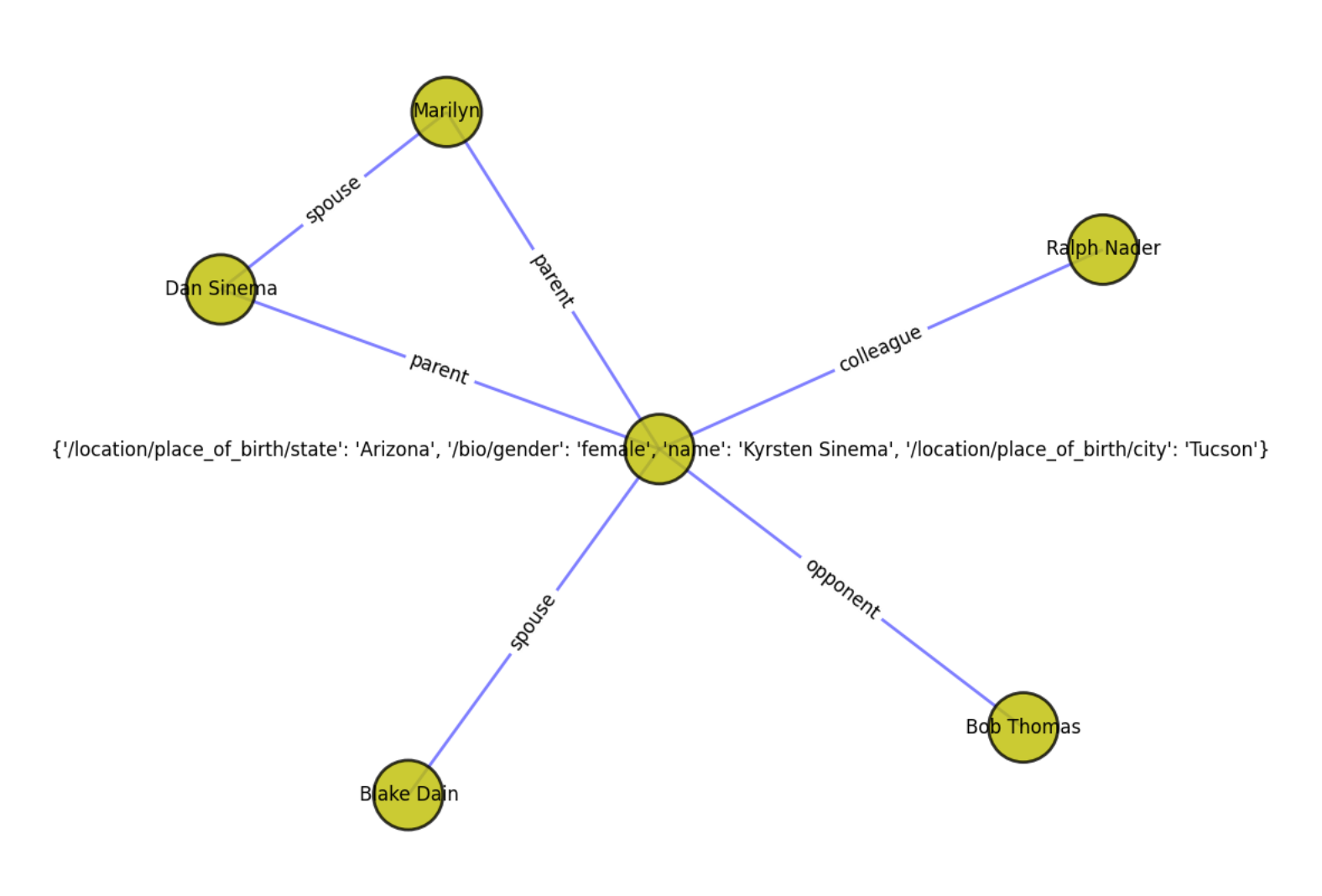}
    \caption{Person Ontology Graph}
    \label{fig:ontology_graph}
\end{figure}

As shown in \ref{tab:linkprediction}, Position Aware Graph Neural Network \cite{you2019position} performs much better than Graph Convolutional Networks \cite{schlichtkrull2018modeling} on both the UDBMS and Elected Representatives datasets.

The Ontology Graph populated by us, parts of which are shown in \ref{fig:ontology_graph}, can be used to improve search, natural language based question answering, and reasoning systems. Further, the graph data can be exported as RDF and PPI formats, and used as a dataset for Link Prediction experiments.

\section{Conclusion}
We introduced a personal data ontology with 36 Entity Types (PDET) and 9 relations (PDER) and annotated unstructured documents from wikipedia using rule based annotators known as SystemT. We then showed improvements to state of the art models for Entity Classification and Relation Extraction. Finally we showed the implementation of a personal data ontology graph population pipeline, incorporating these two neural models along with a Graph Neural Network for Link Prediction.


\bibliographystyle{aaai}
\bibliography{aaai}
\end{document}